\documentclass[lettersize,journal]{IEEEtran}
\usepackage{amsmath,amsfonts}
\usepackage{algorithmic}
\usepackage{algorithm}
\usepackage{array}
\usepackage{textcomp}
\usepackage{stfloats}
\usepackage{url}
\usepackage{verbatim}
\usepackage{graphicx}
\usepackage{soul}
\usepackage{cite}
\usepackage{caption}
\usepackage{subcaption}
\usepackage{hyperref}
\usepackage{xcolor}
\begin{document}

\title{Synthesis and Execution of Communicative Robotic Movements with Generative Adversarial Networks}

\author{
\IEEEauthorblockN{Luca  Garello\IEEEauthorrefmark{1}\IEEEauthorrefmark{2}\IEEEauthorrefmark{4}\IEEEauthorrefmark{6} and Linda Lastrico\IEEEauthorrefmark{1}\IEEEauthorrefmark{2}\IEEEauthorrefmark{4}, Alessandra Sciutti\IEEEauthorrefmark{3},\\ Nicoletta Noceti\IEEEauthorrefmark{2}\IEEEauthorrefmark{6}, Fulvio Mastrogiovanni\IEEEauthorrefmark{2}, Francesco Rea\IEEEauthorrefmark{4}}\\

\IEEEauthorblockA{\IEEEauthorrefmark{6}{MaLGa-DIBRIS, University of Genoa\\}}
\IEEEauthorblockA{\IEEEauthorrefmark{4}Robotics, Brain and Cognitive Science Department (RBCS),\\ Italian Institute of Technology, Genoa, Italy\\}
\IEEEauthorblockA{\IEEEauthorrefmark{2}Department of Informatics, Bioengineering, Robotics, and Systems Engineering (DIBRIS),\\ University of Genoa, Italy\\}
\IEEEauthorblockA{\IEEEauthorrefmark{3}Cognitive Architecture for Collaborative Technologies Unit (CONTACT),\\ Italian Institute of Technology, Genoa, Italy\\}
\IEEEauthorblockA{\IEEEauthorrefmark{1}These authors contributed equally to this work, email: luca.garello@iit.it, linda.lastrico@iit.it \\}
}

\markboth{Special Issue in IEEE Transactions on Cognitive and Developmental Systems (TCDS)}%
{Shell \MakeLowercase{\textit{et al.}}: A Sample Article Using IEEEtran.cls for IEEE Journals}

\maketitle

\begin{abstract}
Object manipulation is a natural activity we perform every day. How humans handle objects can communicate not only the willfulness of the acting, or key aspects of the context where we operate, but also the properties of the objects involved, without any need for explicit verbal description. Since human intelligence comprises the ability to read the context, allowing robots to perform actions that intuitively convey this kind of information would greatly facilitate collaboration. \\
In this work, we focus on how to transfer on two different robotic platforms the same kinematics modulation that humans adopt when manipulating delicate objects, aiming to endow robots with the capability to show carefulness in their movements. We choose to modulate the velocity profile adopted by the robots' end-effector, inspired by what humans do when transporting objects with different characteristics. We exploit a novel Generative Adversarial Network architecture, trained with human kinematics examples, to generalize over them and generate new and meaningful velocity profiles, either associated with careful or not careful attitudes. This approach would allow next generation robots to select the most appropriate style of movement, depending on the perceived context, and autonomously generate their motor action execution.
\end{abstract}


\section{Introduction}
\label{sec:intro}

When it comes to information exchanging, humans rely significantly on non-verbal communication.
This can be explicit and deliberate, as in the case of a pointing gesture, or implicit, as in the case of information encoded in body position, gaze direction, or attitudinal stance while performing a certain gesture \cite{Sandini2018}. 
Such natural, implicit information may be purposefully accentuated in a social environment to make our acts or future intentions more accessible to our partner \cite{signaling}. It would be useful if robots could master similar communicative skills during collaboration, adjusting their goal-oriented activities to make them more intuitively readable to the human partner \cite{legibility}. In particular, referring to the specific context where the robot may need to manipulate objects together with a human partner, it would increase the safety and efficiency of the task if the robot was able to modulate its motor actions according to the properties of the object. Indeed, when we perform an action, even though our movement is primarily goal-oriented and the kinematics optimized to accomplish the movement effectively, it can anyhow reveal to the observers additional information about the context of the action or the characteristics of the objects involved.\\
In item manipulation, the arm motion is influenced by the object weight \cite{weightFlanagan, velWeight}, and people, already from infancy, can instinctively predict the handled weight just by watching others lift it \cite{sciutti:weightChildren}. Another factor that influences how we handle an object is its fragility, that is if it requires particular handling: people deliberately adjust their motor plan to maintain the object's integrity. Previous studies have shown that careful handling has a significant impact on human transportation behavior, to the point where low and high degrees of carefulness could be detected automatically observing the transport motions \cite{carefulnessBillard, icsrLinda}. 
According to such works, the velocity profile of the executed motions can provide useful information about the qualities of handled items. Aware of these findings, in a previous work, we focused on how to create synthetic velocity profiles, suitable for different object characteristics, using Generative Adversarial Networks (GANs) \cite{ICDL_GAN}. The core idea was that, after an adequate training with movements acquired during the manipulation of objects by humans, GANs could allow robots to generalize and autonomously generate sufficiently communicative movements, belonging to the desired class, yet without having to exactly copying human ones. GANs have been commonly used in a number of domains of research, especially in the computer vision one \cite{gan_cv}, since they can create novel realistic data after being trained on a set of real samples. However, their application in the generation of multivariate time series, as human motion kinematics in our case, has been explored only to a limited extent \cite{GAN,mogren2016c,b3}. In particular, there are few examples of GANs being used to generate motion in the context of human interaction \cite{ishiguro,yang2019appgan,butepage2020imitating}. \\
In this work, $(i)$ we present a novel conditional generative architecture for generating velocity profiles associated with the manipulation of objects with different properties; in particular, the transportation of glasses which required Careful (C) or not (Not Careful, NC) handling, given the presence or absence of water inside them. The choice to use a conditional network could allow, in the future, to generate intermediate and completely new data with respect to the two classes used to train the network, whose communicative effectiveness, at the time of movement execution, remains to be evaluated. Moreover, $(ii)$, in this study we focus on the control aspects, transferring the planned movements on two different robots: iCub humanoid robot \cite{iCub} and Baxter robot \cite{fitzgerald2013developing}. These two platforms show remarkable differences in their appearance and dimension, in the distribution of the degrees of freedom (kinematic chain), in typologies of actuators and control strategies; however, they are both provided with two arms resembling a humanoid appearance and offer the possibility of transporting objects with their end-effectors. For these reasons, they are particularly suitable for testing the replicability of our approach, comprising both motor action generation and motor control and its generalization potential. We evaluate the performances of the two robots in following the desired trajectories while modulating the velocity of their end-effector according to the selected velocity profile. In the future, we plan to further exploit the approach presented in this study and carry out interaction experiments involving the two robots. This would allow us to assess if the reproduced movements are communicative enough of the object properties, if the reactions elicited depends on the modulation of the velocity profile and finally how the different embodiment of the robots influence the interaction.

\section{Why a generative approach}

\begin{figure}[t]

\begin{subfigure}[b]{\columnwidth}
\centering
\includegraphics[width=0.9\columnwidth]{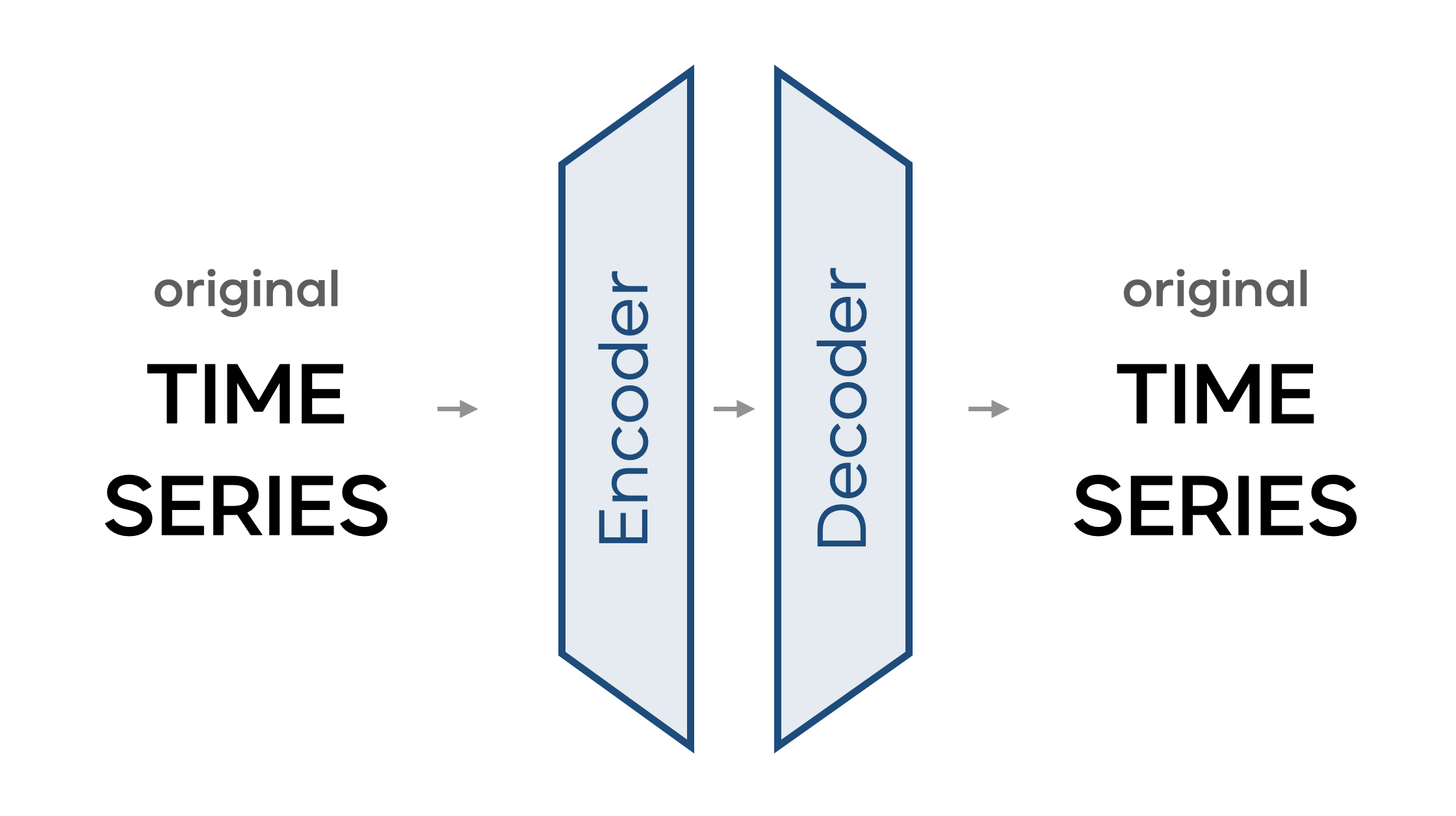}
\caption{Autoencoder Pretraining}
\label{fig:Pretraining}
\end{subfigure}

\begin{subfigure}[b]{\columnwidth}
\centering
\includegraphics[width=0.9\columnwidth]{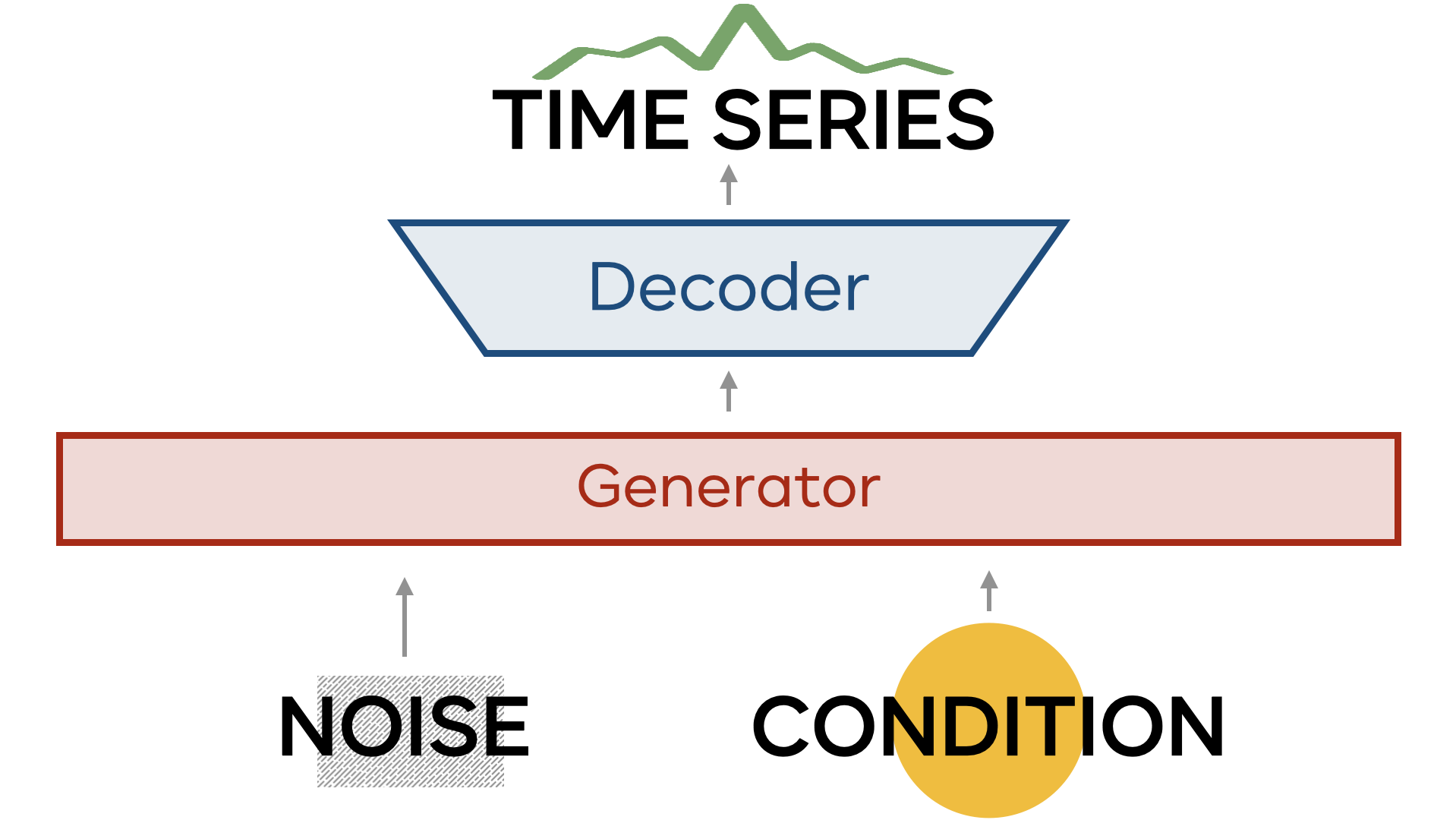}
\caption{Inference}
\label{fig:Generation}
\end{subfigure}
\caption{\textbf{Proposed architecture:} in (\ref{fig:Pretraining}) is represented how is performed the training of the encoder/decoder network, to represent the time series in the embedded space. In (\ref{fig:Generation}), are shown the steps for synthesizing a new velocity profile: our model takes as input random noise and a conditioning label, then the generator outputs in an embedding space that represents the characteristics of the signal we want to generate. Finally, the previously trained decoder translates the desired embedded properties into a time series}
\label{fig:Architecture}
\end{figure}

A natural interaction with robots requires them to have a rich motor repertoire. This because stereotyped and repetitive movements can make the interaction less natural and fluid.
If we take as an example any action performed with an arm, it can be executed with many trajectories and different velocity profiles. Although in the literature there are already works that have attempted to generate movements with different trajectories \cite{ishiguro,yang2019appgan,butepage2020imitating}, to the best of our knowledge there are no studies on generating movements purposely mapping the velocity profiles, especially if the velocity modulation is aimed to implicitly describe an emotional state or the properties of the manipulated object.
One possible approach, to generate communicative movements, would be to perform pre-recorded human movements on the robots, however this solution would not make the robots truly ``autonomous". For this reason, we tried to fill this gap by developing a generative neural network capable of synthesizing new velocity profiles that are descriptive of the properties of an object manipulated by the robot.
The first goal of our generative model is to obtain synthetic data which preserve the temporal dynamics of the original time-series. To produce velocity profiles that reflect the two original classes C and NC, we exploit a two-steps training process. At first, we train an autoencoder to learn embedded representations of the original time-series. Then, such embeddings are used to train our cGAN, which generates new synthetic embeddings. The generator can be conditioned to output embeddings belonging to one of the two classes by giving as input a vector of labels and a random time-series noise (see Figure \ref{fig:Architecture}). Finally, a decoder is employed to reconstruct the time-series from the generated embedded representations.
The main advantage of a generator which directly outputs in the embedded space is that in this way we encourage the model to focus on the relevant features of the dataset rather than on the samples themselves \cite{GAN}. 

\section{Generative problem statement}
Suppose to have a distribution $p_A$ of time series that describes the process $A$ and another distribution $p_B$ that describes the process $B$.
We want to learn a Generator distribution $p_g(z|y)$, where $z$ is random noise from a normal distribution, that is conditioned by some extra information $y$, for example a numerical label describing the target class we want to generate.
Then $p_g(z|y)$ can be learnt by a generator $G(z,y;\theta_g)$ whose output $x$ should lie in $p_A$ or $p_B$ based on the condition $y$.
This learning process is supervised by a Discriminator $D(x,y;\theta_d)$, which outputs a single scalar representing the probability that $x$ came from training data rather than $p_g(z|y)$.\\
We propose a conditional model for the generation of time series. Our model consists of four network components: encoder, decoder, sequence generator and sequence discriminator (see Figure \ref{fig:Architecture}). The training of the first two components allows to learn encoded representations of the time series in the dataset. The  adversarial network (\textit{generator} + \textit{discriminator}) is exploited to learn to generated embedded representations in the latent space. These synthetic embeddings can be then used by the decoder to reconstruct plausible time series.

\begin{figure*}[]
\centering
  \begin{subfigure}[b]{0.45\textwidth}
    \centering
    \includegraphics[width=1\linewidth]{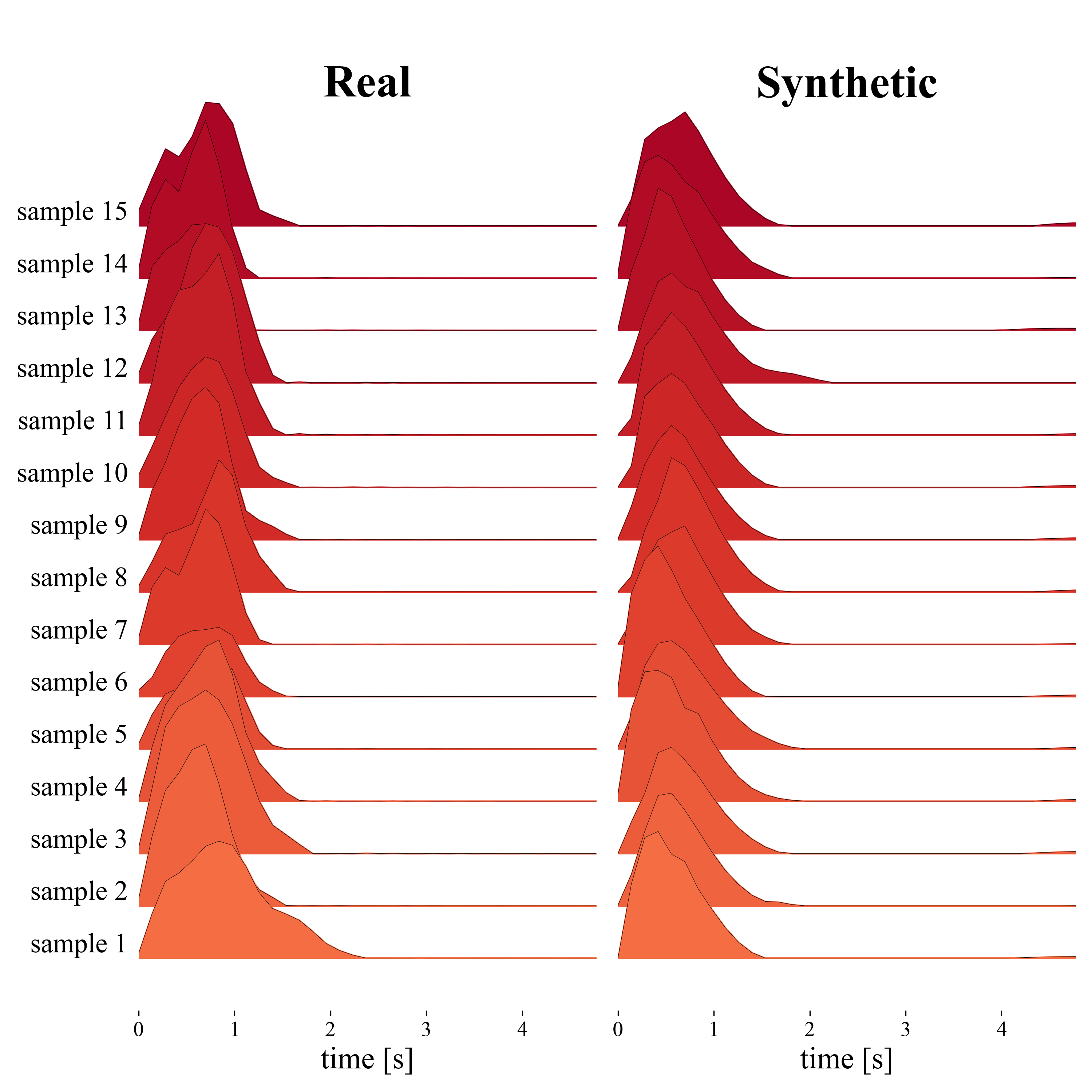}
    \caption{Not Careful}
    \label{fig:series_NC}
  \end{subfigure}
  \hspace{1em}%
  \begin{subfigure}[b]{0.45\textwidth}
    \centering
    \includegraphics[width=1\linewidth]{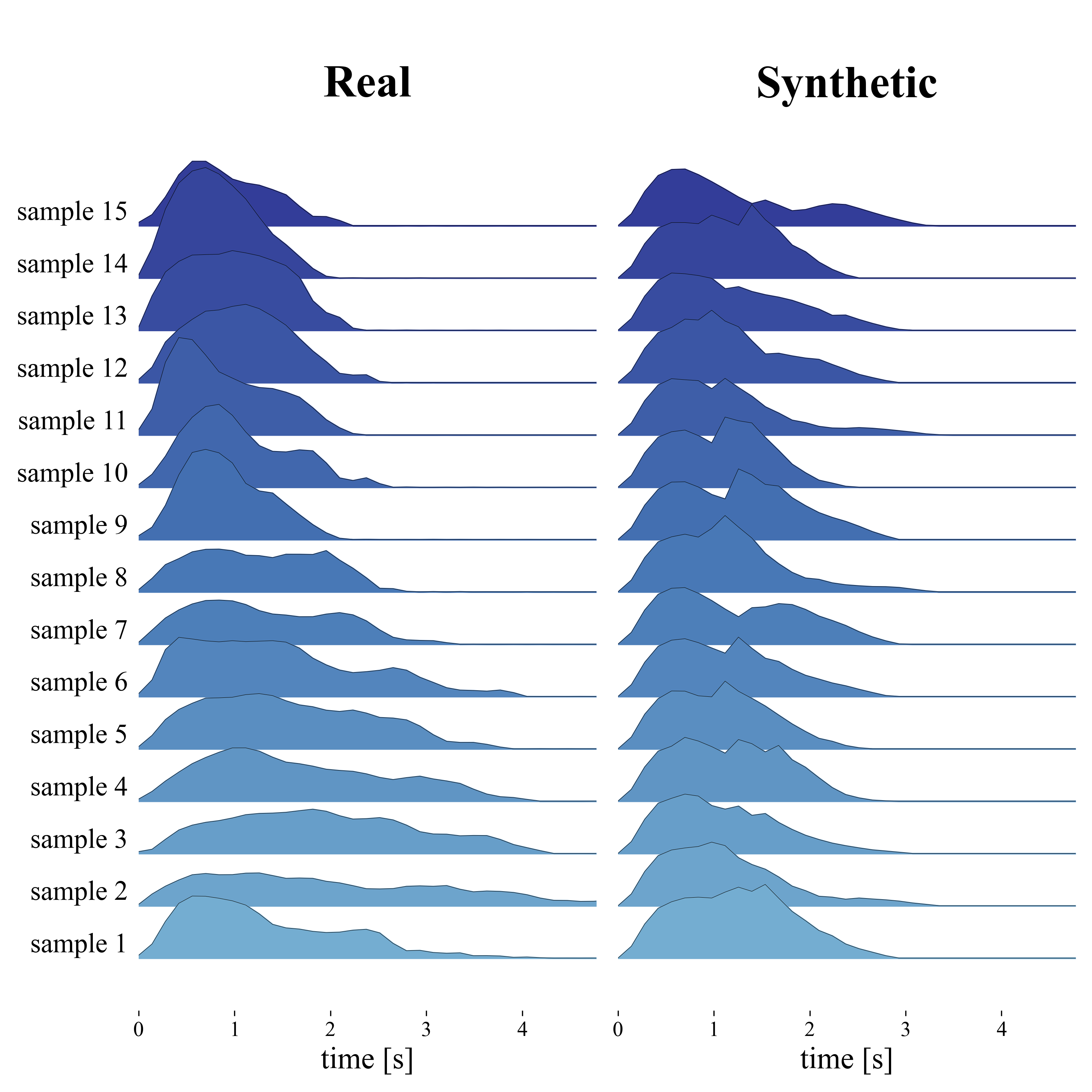}
    \caption{Careful}
    \label{fig:series_C}
  \end{subfigure}
  \caption{\textbf{Comparison between real and synthetic data:} representation of a subset of the original velocity profiles that constitute the training data (real) and of the time series synthesized by the GAN, for both the Not Careful (\ref{fig:series_NC}) and Careful (\ref{fig:series_C}) classes. Please note that there is no one-to-one correspondence between each Real and Synthetic sample: they are randomly selected to give a qualitative overview of the original and synthetic dataset}
  \label{fig:series}
\end{figure*}

\begin{figure}
    \centering
    \includegraphics[width=\columnwidth]{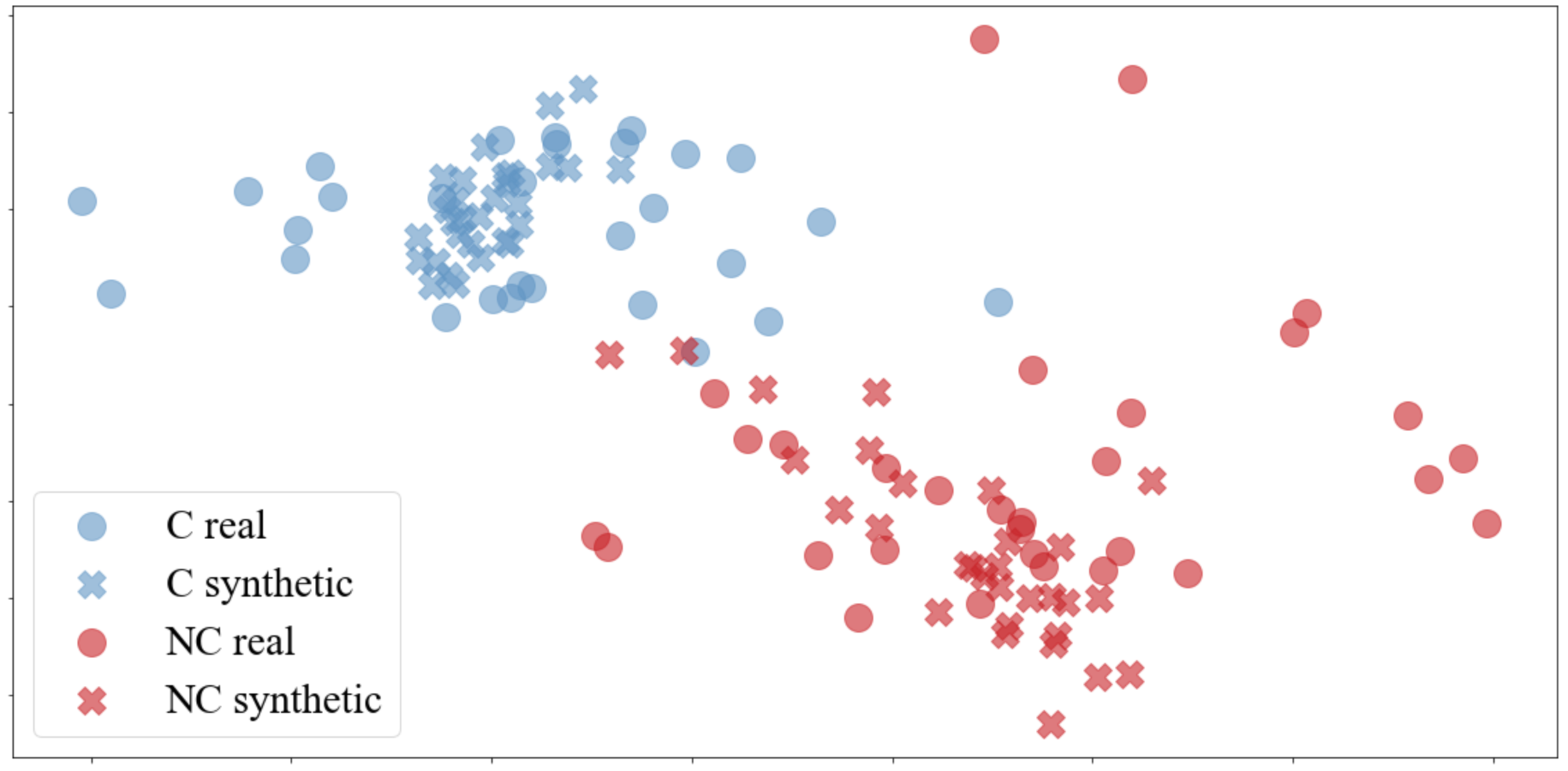}
    \caption{\textbf{PCA}: principal component analysis of the original and synthetic data distributions.
    The synthetic samples are close to the target classes but without overlaps, meaning that they are not copying the originals}
    \label{fig:PCA}
\end{figure}

\subsection{Encoder/Decoder networks}
The \textit{encoder/decoder} networks are jointly trained as an \textit{autoencoder} before starting the adversarial training of \textit{generator} and the \textit{discriminator}.
The \textit{encoder/decoder} (Figure \ref{fig:Pretraining}) provide mappings between the original time series and latent space, enabling then the adversarial network to learn about the data underlying temporal dynamics through lower-dimensional learning.  The goal is to learn an embedded representation of the original data. This compressed representation, together with the output of the discriminator, is used as a supervision signal during training. For our work we implemented a simple LSTM with one encoder layer and one decoder layer. 
\subsection{Generator/Discriminator networks}
After the training of the \textit{encoder/decoder} networks, the \textit{generator} and the \textit{discriminator} are trained.
In the generator the prior input noise $p_g(z)$ and the conditioning label $y$ are combined in a joint hidden representation. The label is embedded into a vector of the same size as $z$ and then multiplied for the $z$ vector. This input is fed to the generator which is composed of 1 LSTM layer, and the generator outputs an embedding.
This compressed representation is exploited for computing the supervised loss as shown in \ref{loss}.

\begin{equation}
\arg \min_G \max_D V(D,G) = \mathcal{L}_{GAN}(G,D)+\lambda \mathcal{L}_{L2}(G)
\label{loss}
\end{equation}
where:
\begin{equation}
\mathcal{L}_{GAN} (G, D) = \mathbb{E}_{x,y} [\log D(x, y)]+ \\
\mathbb{E}_{x,z} [\log(1 - D(x, G(z, y))]
\end{equation}
and:
\begin{equation}
\mathcal{L}_{L1}(G)=\mathbb{E}_{x,y,z}[\left \|  x - G(z, y) \right \|_{2}]
\end{equation}

The loss function is composed of two terms, \textit{(i)} the output of the discriminator,  that needs to be fooled, and \textit{(ii)} the l2 norm between the embeddings generated and those obtained by encoding the original data through the pretrained encoder.
For stability reasons the L2 loss can be modulated by a constant factor lambda ($\lambda$), in our experiments equal to $100$.
At inference time the embeddings produced by the \textit{generator} can be finally reconstructed by the \textit{decoder}.
See Figure \ref{fig:Generation} for reference.
Some of the velocity profiles generated by our model are shown in \ref{fig:series} while \ref{fig:PCA} offers a 2D visualization of how the synthetic samples match the original data distributions. 

\subsection{Dataset description}
 As in our previous work \cite{ICDL_GAN}, we focused on the generation of velocity profiles associated with the transportation of objects with different properties, exploiting the same dataset. Details about the dataset structure and acquisition can be found in \cite{ICDL_GAN} and \cite{HFR}. For training the GAN, we considered 1001 total transport movements, recorded with a sampling rate of 22 Hz with active infrared markers placed on the participants right hand. The objects involved were four glasses, identical in shape and appearance, but differing for the weight (light: 167 gr; heavy: 667 gr) and content: two of them were completely filled with water, necessitating extra caution to be moved. The experimental setting inherently allowed for a significant degree of heterogeneity in the movements performed: trajectories may be directed to the left or right side, to lower or higher shelves, and the motion might be abductive or adductive. The trials belonged to different categories depending of the properties of the objects: for the NC class, 248 gestures involved the light glass and 251 the heavy one; regarding the glasses full of water (Careful class) 254 transport motions were carried out with the light glass and 248 with the heavy one. In training the GAN, we only considered as feature the carefulness required, while the weight difference counted as added variability in the dataset.\\

\section{Robots' movements generation} \label{robot:movegen}
The second part of our study focuses on the possibility of replicating the desired velocity profiles with the robots end-effector. Independently from the nature of the velocity profile (either synthetic or recorded from human examples), we want the robots to accurately reproduce the desired velocity profiles at the end-effector, maintaining the most communicative characteristics of the movement.
We considered two robots, the humanoid robot iCub \cite{iCub} and the collaborative robot Baxter \cite{fitzgerald2013developing}, in order to perform rigorous analysis independently of the kinematics and control of the robots.
Although these two robotic platforms present significant differences in terms of appearance and kinematics (e.g. spatial configuration of the Degrees Of Freedom) they both share the ability to interact with humans and to manipulate objects.
To assess the ability of the robots in following a generic velocity profile, we randomly selected three examples among the Careful class (\textit{ProfileC1, ProfileC2, ProfileC3}), and three among the Not Careful class (\textit{ProfileNC1, ProfileNC2, ProfileNC3}). We also decided to evaluate the movement executions along three different planes: frontal, sagittal and oblique (intermediate between the other two). Due to the different kinematic configurations of the robots, their range of motion and workspace differed. All trajectories generated for Baxter covered a path of $60$ $cm$, while those performed by iCub were about $30$ $cm$, depending on the plane on which the movement occurred. 
Finally, for each of the six selected velocity profiles and of the three trajectories, we executed the same movement ten times, to have a measure of the robots' consistency in repeating the actions. See \href{https://youtube.com/playlist?list=PL9sy5y8WKC4kX4AMGCdXYneK6-VqaR4M5}{robotMovements-videos}\footnote{Movements execution on Baxter and iCub robots \url{https://youtube.com/playlist?list=PL9sy5y8WKC4kX4AMGCdXYneK6-VqaR4M5}} for a clear view of the generated movements. For both the robots, we saved the end-effector cartesian position during the trajectory execution for subsequent analyses. In particular, we computed the tangential velocity magnitude of the end-effector to compare it with the one planned by the controllers. 

\subsection{Baxter robot}
We decided to control the Baxter Robot with MoveIt!, the standard ROS motion planning framework \cite{moveit}. This choice was driven by the intention to use a framework familiar to all those who use Baxter without necessarily installing additional software or possessing advanced knowledge of control theory.
We first remapped the selected velocity profiles to fit the desired movement length, we kept constant the temporal duration of the action while modulating the magnitude of the velocity. Then the desired trajectory was divided into \textit{n} Cartesian points equally spaced by 0.01 m. The choice of using a fixed spatial step, while intervening on the temporal execution was driven by the MoveIt! controller implementation.
At this point MoveIt! was asked to plan a trajectory defined by these points. Such trajectory is defined a sequence of poses in the joints space. During the planning we imposed to each spatial point the time with which it was supposed to be reached. In this way it was possible to define the cartesian velocity of each spatial step.

\subsection{iCub robot}
Similarly to Baxter we controlled the iCub robot with its default ``Cartesian Controller" developed by Pattacini et al. \cite{pattacini2010experimental}. 
Also on iCub we have implemented movements that preserved the original duration. On iCub the end-effector controller required to fix the execution time of each sub-step, varying only the Euclidean distance between them.

\begin{figure*}[]
\centering
  \begin{subfigure}[b]{0.46\textwidth}
    \centering
    \includegraphics[width=1\linewidth]{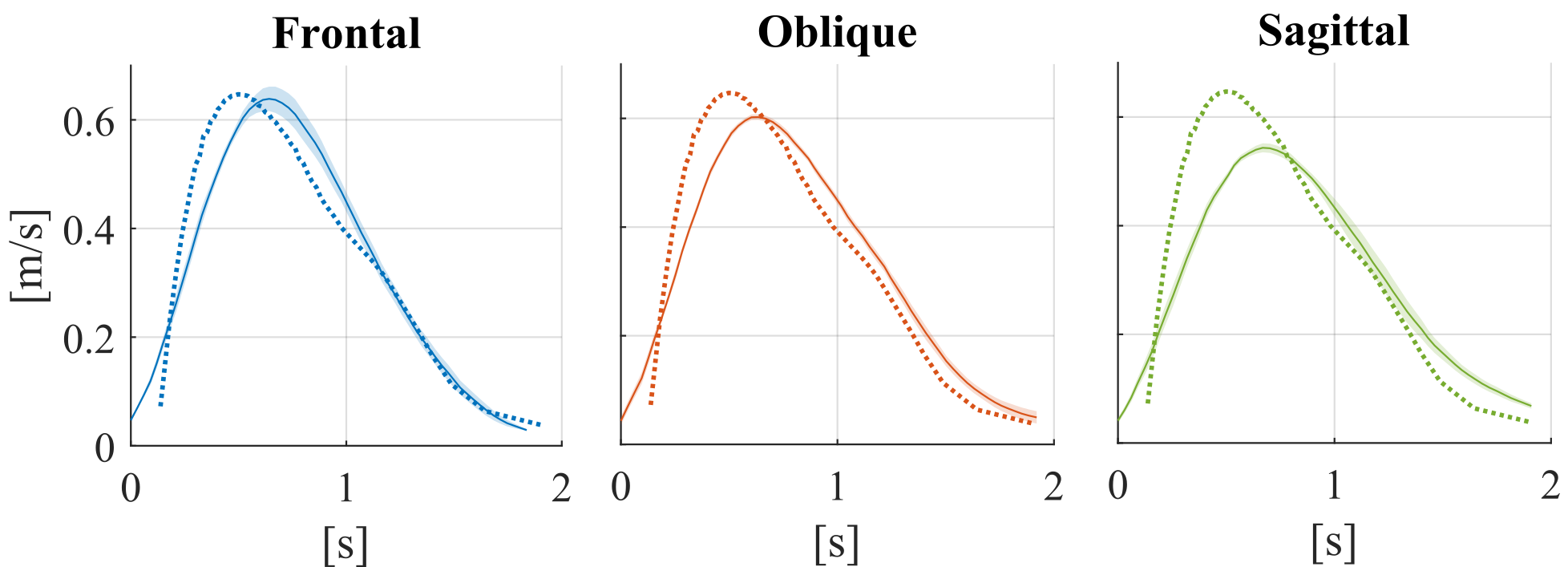}
    \caption{Not Careful - ProfileNC2}
    \label{fig:baxter_vel_NC}
  \end{subfigure}
  \hspace{1em}%
  \begin{subfigure}[b]{0.46\textwidth}
    \centering
    \includegraphics[width=1\linewidth]{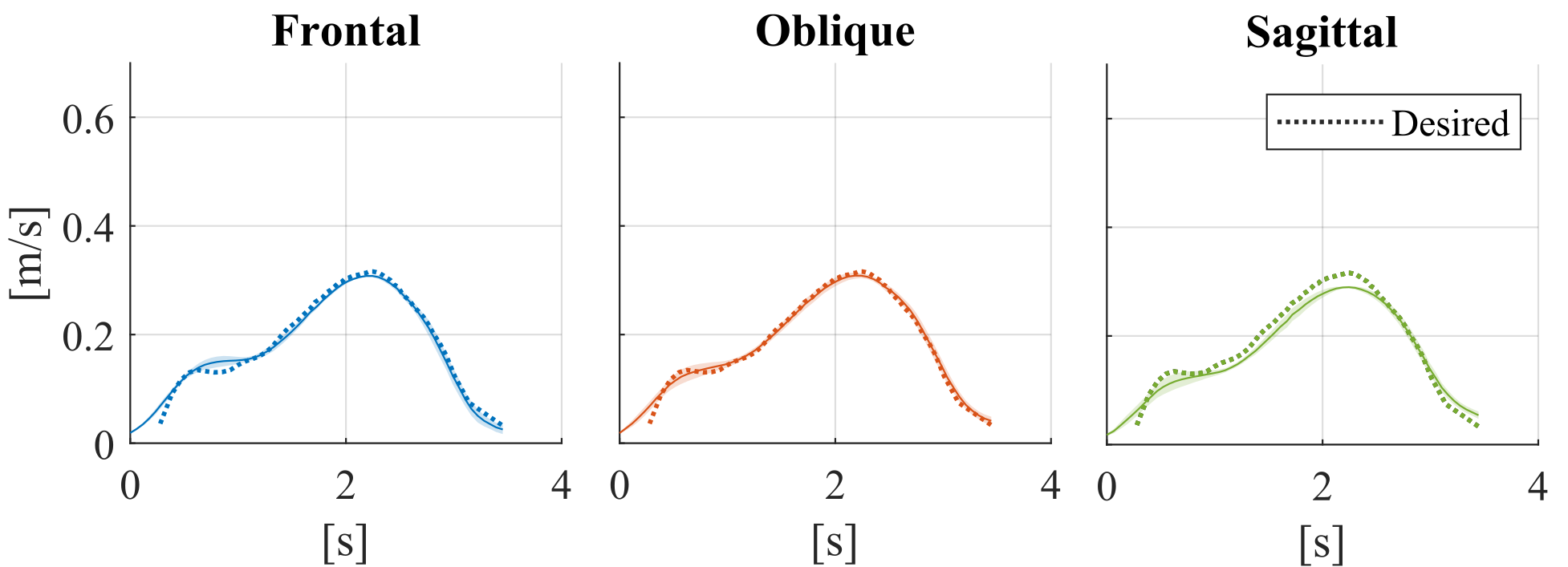}
    \caption{Careful - ProfileC1}
    \label{fig:baxter_vel_C}
  \end{subfigure}
  \caption{\textbf{Baxter executed velocity profiles:} The represented profiles are respectively ProfileNC2 for the Not Careful class (\ref{fig:baxter_vel_NC}) and ProfileC1 for the Careful one (\ref{fig:baxter_vel_C}). Ten repetitions of such profiles are executed along three different movement directions: frontal, sagittal and oblique. The mean of the velocity executions is a solid line, while the standard deviation is a more transparent area of the same colour. The desired velocity profile is dashed and it is identical in the three directions, since the distance covered by the Baxter robot was the same}
  \label{fig:baxter_velocities}
\end{figure*}
\begin{figure*}[]
\centering
  \begin{subfigure}[b]{0.46\textwidth}
    \centering
    \includegraphics[width=1\linewidth]{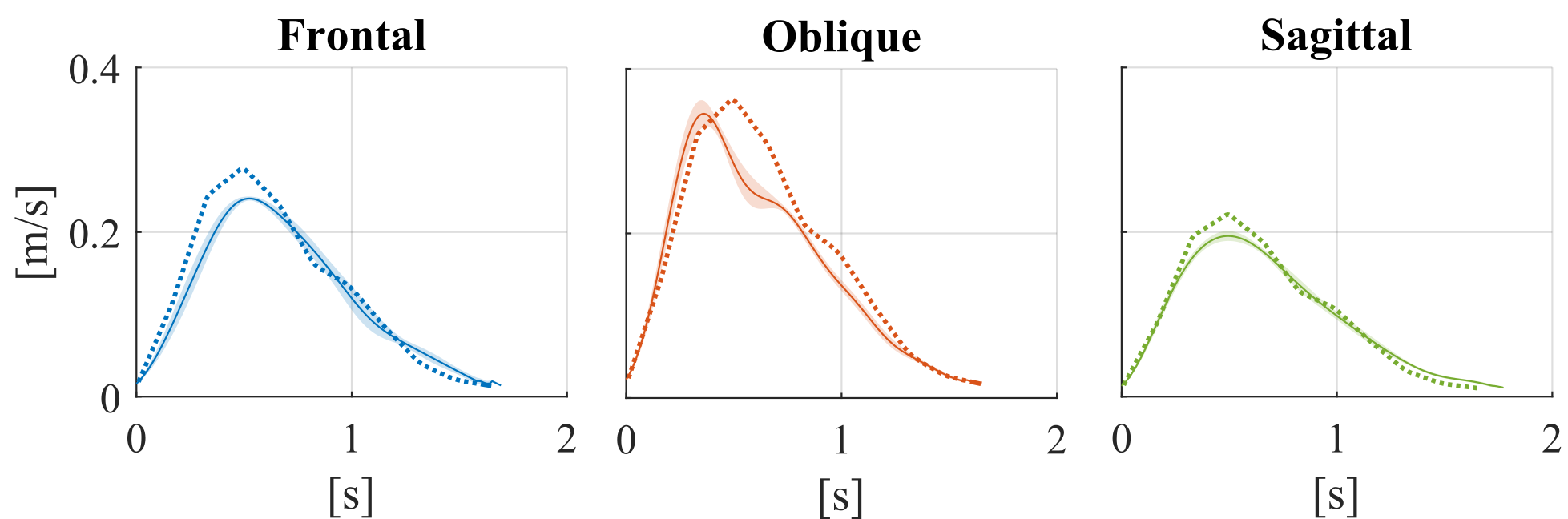}
    \caption{Not Careful - ProfileNC3}
    \label{fig:icub_vel_NC}
  \end{subfigure}
  \hspace{1em}%
  \begin{subfigure}[b]{0.46\textwidth}
    \centering
    \includegraphics[width=1\linewidth]{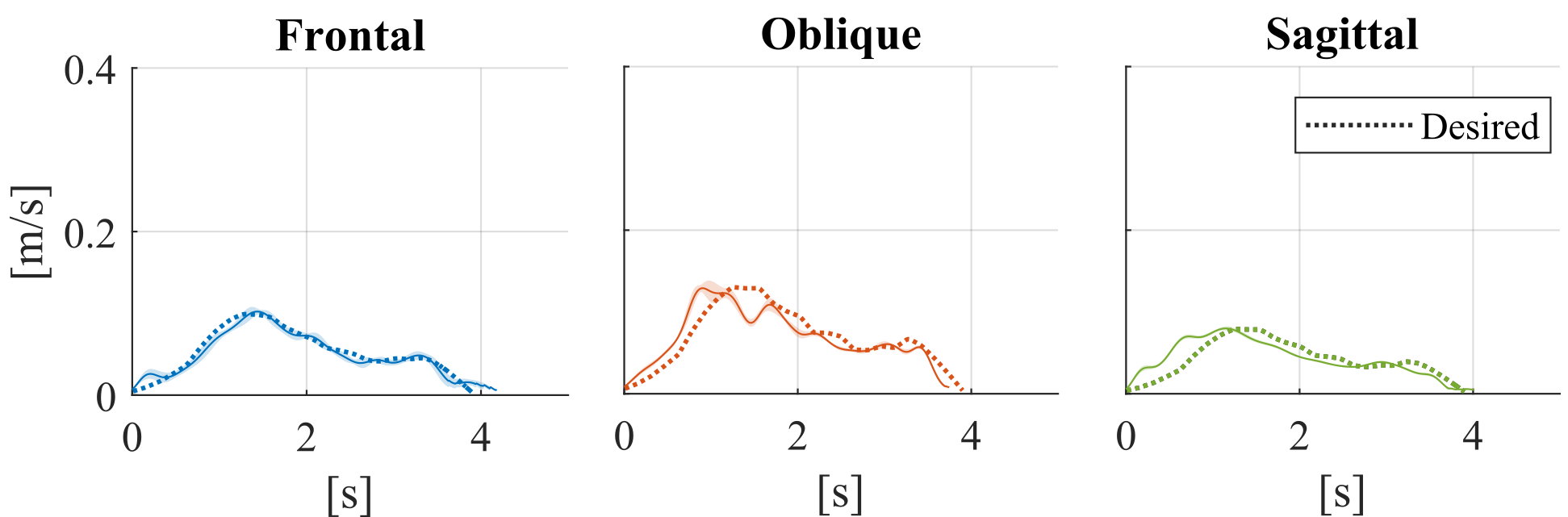}
    \caption{Careful - ProfileC3}
    \label{fig:icub_vel_C}
  \end{subfigure}
  \caption{\textbf{iCub executed velocity profiles:} The represented profiles are respectively ProfileNC3 for the Not Careful class (\ref{fig:baxter_vel_NC}) and ProfileC3 for the Careful one (\ref{fig:baxter_vel_C}). Ten repetitions of such profiles are executed along three different movement directions: frontal, sagittal and oblique and the graphical conventions are the same of Figure \ref{fig:baxter_velocities}, with the desired velocity profiles dashed}
  \label{fig:icub_velocities}
\end{figure*}
\section{Robots' Movements Assessment} \label{robotresults}
To assess the performances of the controller  in replicating our reference velocity profiles, we compared the executed movements with the desired ones. First, in Sec. \ref{results:exec}, we provide a qualitative overview of the performed motions. Then, we compare the maximum velocity amplitude reached with the desired one; this, to investigate whether the robots can reach the same range of velocity that humans apply when manipulating objects, delicate or not (see Sec. \ref{results:magnitude}). Finally, we compute the Pearson correlation between each executed velocity profile and the planned one (see Sec. \ref{results:pearson}).\\
As previously explained in Sec. \ref{robot:movegen}, we randomly selected from our dataset three velocity profiles to exemplify the C class and three for the NC one. We then considered three possible directions for the robots' end-effectors to move: along the frontal, sagittal and transverse plane. Each velocity profile was rescaled according to the selected trajectory, and each movement was repeated ten times, to assess the robot repeatability. This led us to have, for each one of the three directions, ten repetitions of the same movement for each velocity profile considered.
\subsection{Movement execution}
\label{results:exec}
In Figure \ref{fig:baxter_velocities} is shown how the Baxter robot executed, along the three different planes, one among the three Not Careful (\ref{fig:baxter_vel_NC}) and Careful (\ref{fig:baxter_vel_C}) velocity profiles. Even though the movement repetitions along the same direction are comparable, with a very low intra-class variability, some trajectories appear more suitable than others in reproducing the velocity at the end-effector; it can be noticed how, for the Not Careful movements (Figure \ref{fig:baxter_vel_NC}), along the Frontal plane the velocity peak reaches the desired one, while in the Sagittal one it remains lower. The execution of the Careful movements (Figure \ref{fig:baxter_vel_C}) shows no delay with respect to the planned velocity, with the profile peak and shape preserved for all the movement directions.\\
Regarding iCub, in Figure \ref{fig:icub_velocities} are represented two others velocity profiles among the selected ones, different from those represented for Baxter in Figure \ref{fig:baxter_velocities}, one for the NC (\ref{fig:icub_vel_NC}) and one for the C (\ref{fig:icub_vel_C}) class. On this robot, the shape of the not careful profiles is well aligned with the planned movement, and the velocity amplitude almost meets the desired one. Considering Figure \ref{fig:icub_vel_C}, a small delay between the planned and executed movements can be noticed for the Oblique and Sagittal directions.

\subsection{Maximum velocity}
\label{results:magnitude}
\begin{figure}
\centering
  \begin{subfigure}[b]{0.48\textwidth}
    \centering
    \includegraphics[width=1\linewidth]{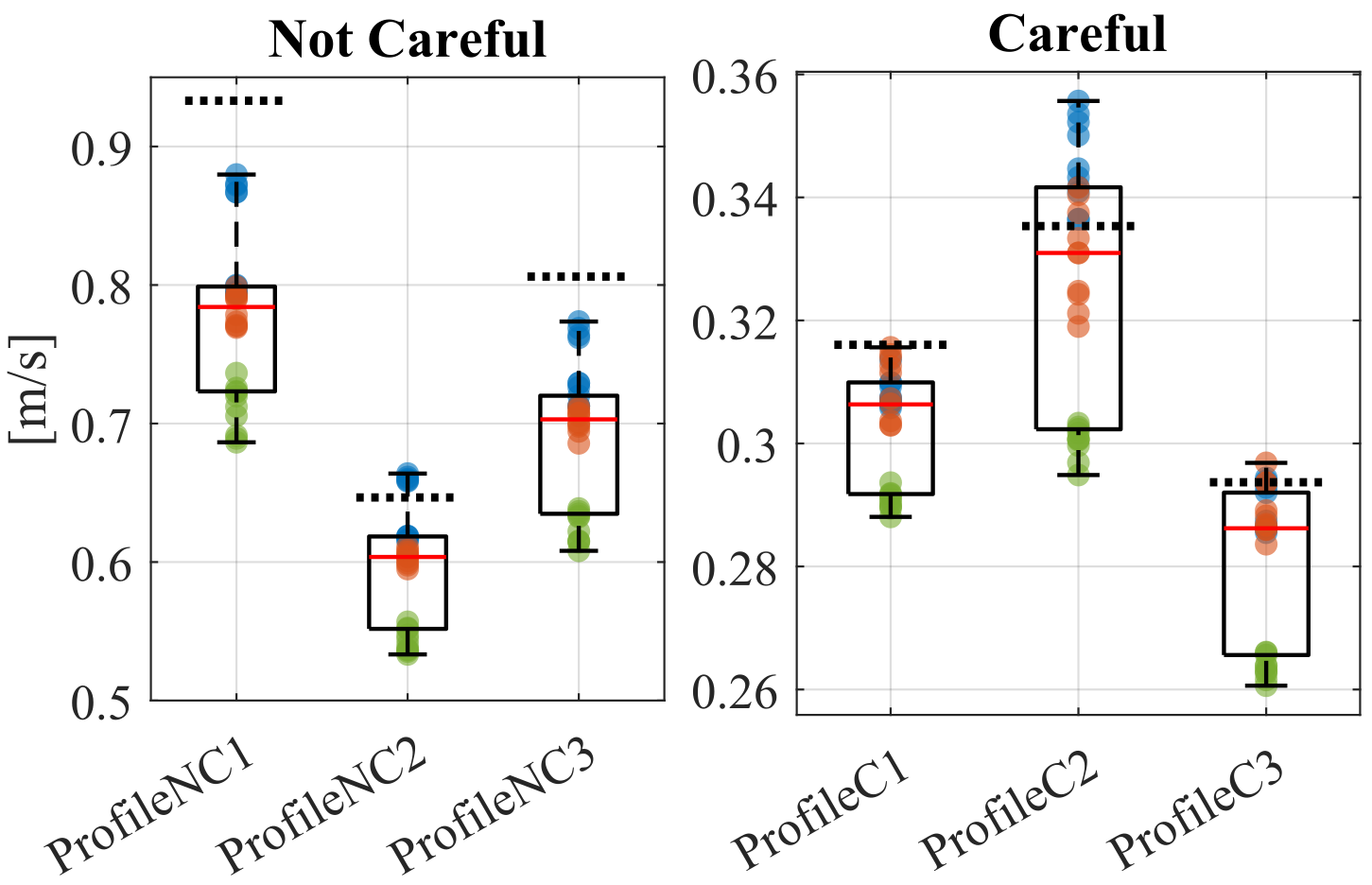}
    \caption{Baxter robot}
    \label{fig:baxter_peaks}
  \end{subfigure}
  \hfill
  \begin{subfigure}[b]{0.48\textwidth}
    \centering
    \includegraphics[width=1\linewidth]{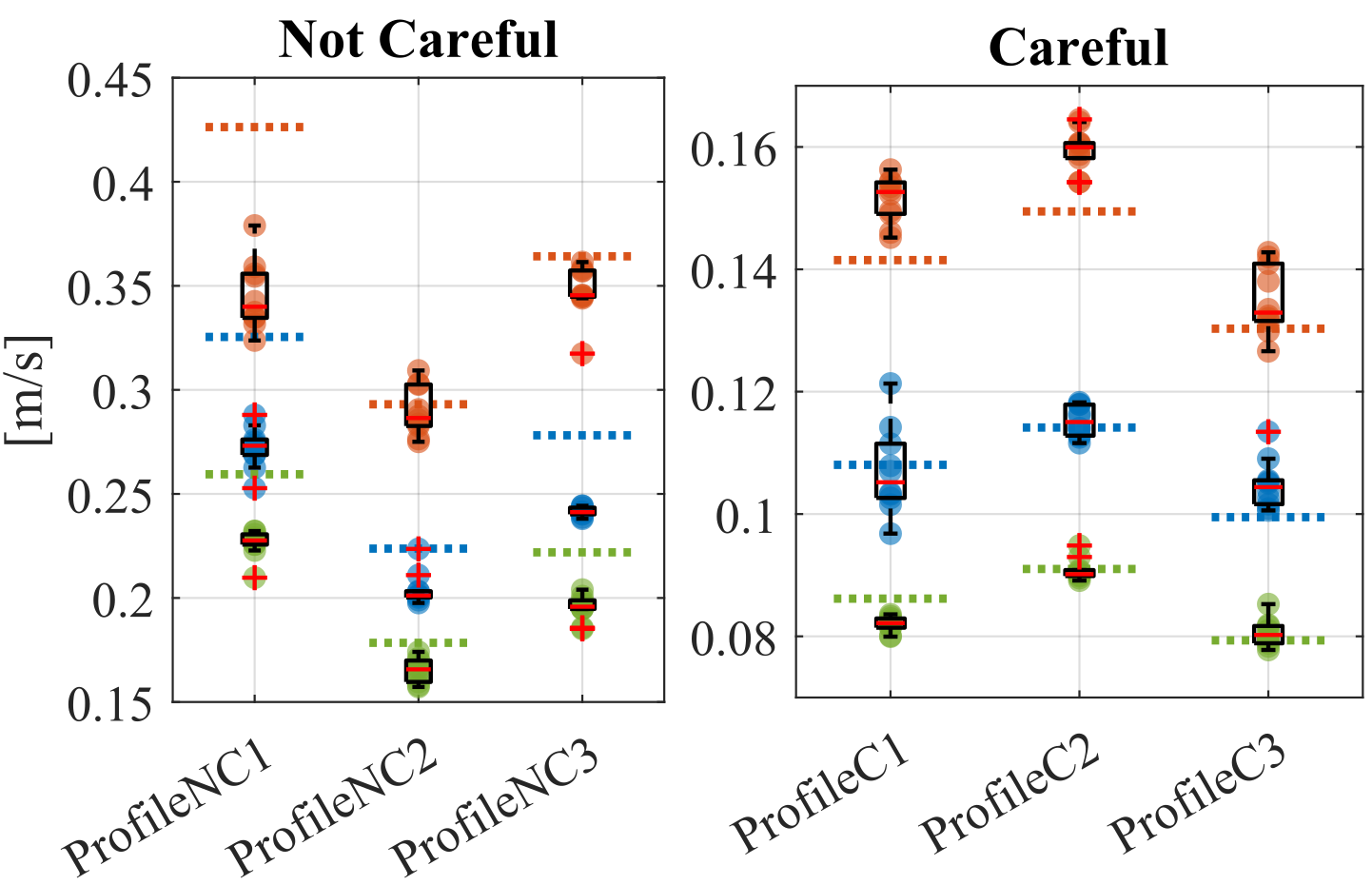}
    \caption{iCub robot}
    \label{fig:icub_peaks}
  \end{subfigure}
  \caption{\textbf{Amplitude of the velocity peaks}: the desired amplitude is marked with a black dashed horizontal line for Baxter (\ref{fig:baxter_peaks}) robot, while it follows the color code for iCub (\ref{fig:icub_peaks}): blue for frontal movements, orange for oblique and green for the sagittal ones. The red lines in the boxplots represent the medians, the black rectangles limit the 25\textsuperscript{th} and 75\textsuperscript{th} percentiles of the executed velocities. Since iCub covered paths of varying length, depending on the direction, we report in Figure \ref{fig:icub_peaks} the boxplot for each plane}
  \label{fig:peaks}
\end{figure}
In Figure \ref{fig:peaks} is represented the maximum velocity reached by the two robots end-effectors together with the planned one. For each example selected for the two classes Not Careful and Careful, is shown the maximum speed reached in every repetition. \\
For the Baxter robot (Figure \ref{fig:baxter_peaks}), the desired maximum speed is indicated by a single black horizontal bar; since the trajectories along the three planes covered the same distance of $60$ $cm$, the associated velocity profiles were rescaled in the same way by the controller. Therefore, the Baxter robot was required to reach the same maximum speed independently from the movement direction. While the distribution of NC movements do not reach the desired speed, for most of the trials, the careful amplitudes are generally comparable with the planned one, with some variability depending on the trial considered. As anticipated by Figure \ref{fig:baxter_velocities}, some trajectories are more suitable than others in reproducing the desired speed, and this reflects also in the maximum peak distributions. Indeed, Baxter reached different maximum speeds depending on the direction: for both the NC and C classes, and independently from the selected velocity profiles, the movements along the sagittal plane, in green, have the lowest maximum speed, followed by the ones in the oblique plane and lastly by the ones in the frontal plane, in blue, which are closest to the desired maximum speed. This reveals a trajectory dependency in the Baxter performances, that should be taken into account when generating new movements.\\
Considering iCub, the path covered along the three planes differed, due to the range of motion of each joint in the kinematic chain: the workspace along the three planes was different. Therefore, the same original velocity profile was rescaled differently for the three movements (see Figure \ref{fig:icub_velocities}, where examples of the desired velocity profiles for the NC and C classes are mapped in the three directions). Figure \ref{fig:icub_peaks} reports the speed distributions with colored dashed lines indicating the desired values; it can be appreciated how, in most of the cases, the distribution associated with each direction of motion is comparable with the corresponding planned values. In particular, the oblique direction in orange required faster movements, while the sagittal one in green the slowest. This was deliberately planned in advance for the iCub robot, according to the trajectory, as confirmed by the dotted reference lines. However, it can be noticed how, for the NC class, iCub executed peaks are systematically lower than the desired ones, with the exception of the oblique trajectories with \textit{ProfileNC2}, that are totally satisfying.

\subsection{Pearson Correlation Coefficient}
\label{results:pearson}

\begin{figure}
\centering
  \begin{subfigure}[b]{0.48\textwidth}
    \centering
    \includegraphics[width=1\linewidth]{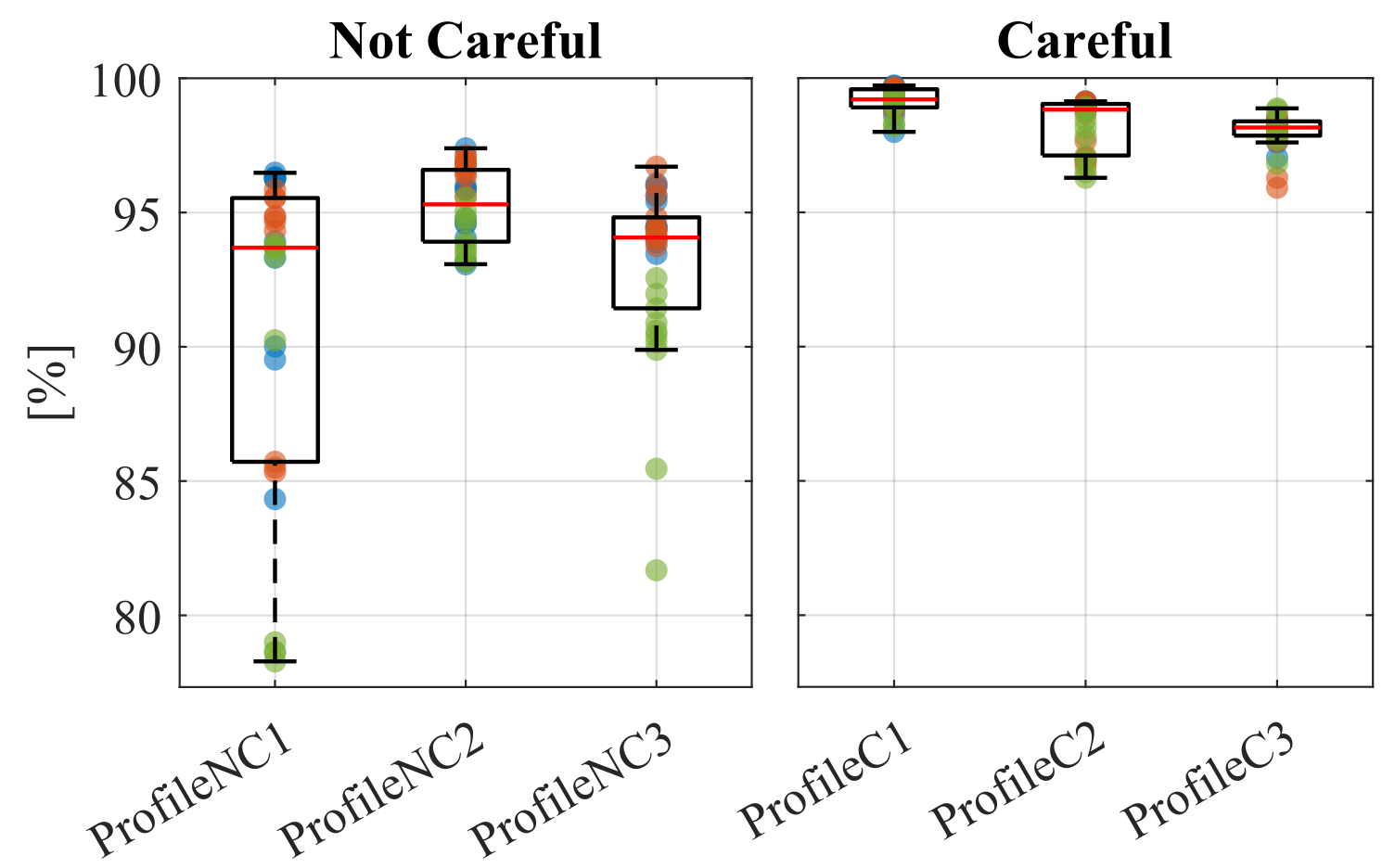}
    \caption{Baxter robot}
    \label{fig:baxter_pears}
  \end{subfigure}
  \begin{subfigure}[b]{0.48\textwidth}
    \centering
    \includegraphics[width=1\linewidth]{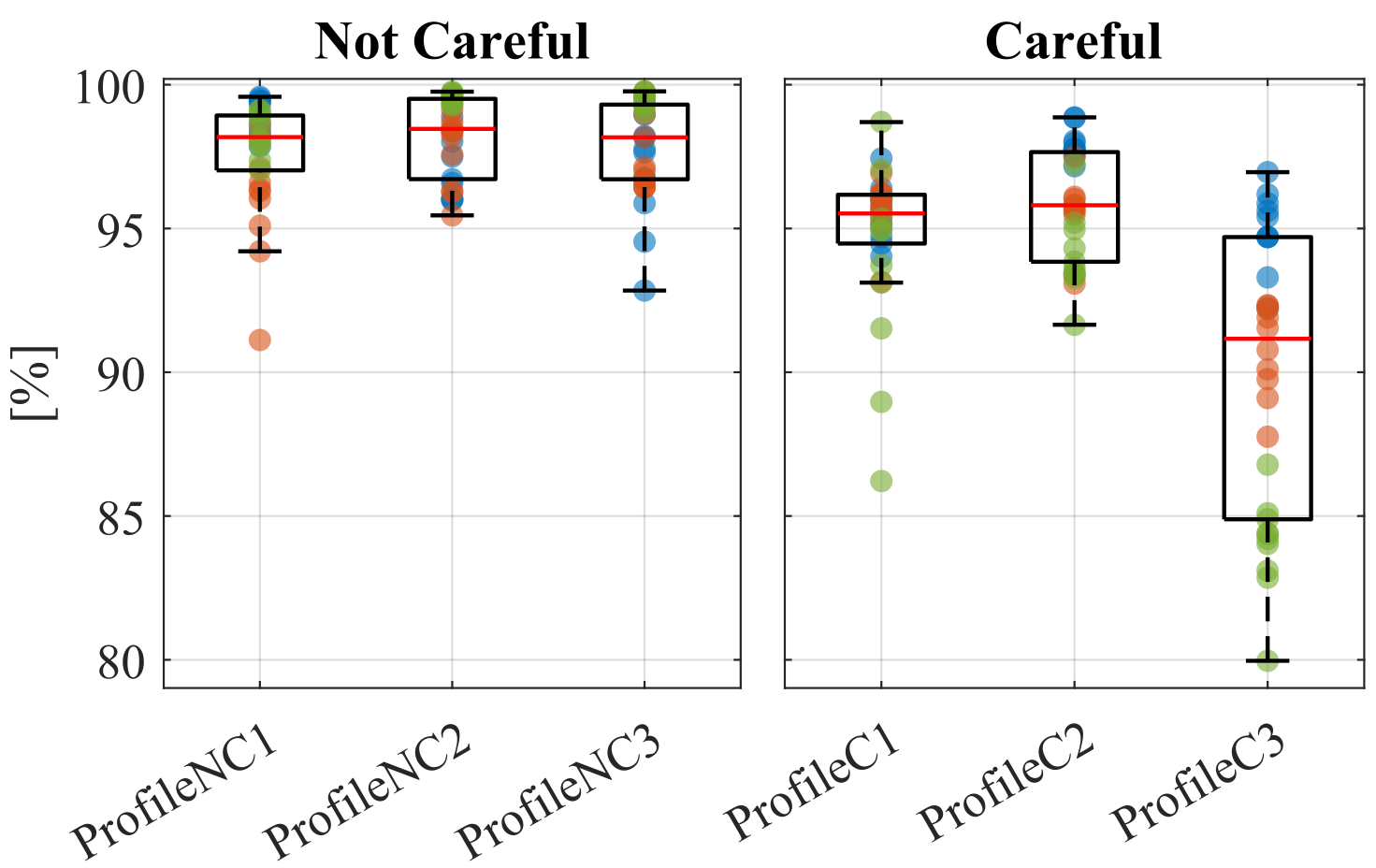}
    \caption{iCub robot}
    \label{fig:icub_pears}
  \end{subfigure}
  \caption{\textbf{Pearson correlation} on Baxter (\ref{fig:baxter_pears}) and iCub (\ref{fig:icub_pears}) robot. For every movement, the correlation has been calculated between the planned and the executed velocity profile. The graphical conventions are the same as in Figure \ref{fig:peaks}, with blue dots for frontal movements, orange for oblique and green for the sagittal ones}
  \label{fig:pearson}
\end{figure}

Finally, in Figure \ref{fig:pearson}, we considered the Pearson correlation coefficient to evaluate how well the executed velocities are related to the planned ones. In Figure \ref{fig:baxter_pears} are represented the correlation coefficients for every trial obtained with Baxter robot. For the Not Careful class, the median correlation settles in the $94-95\%$ range, while in the Careful class is above the $98\%$ in all the three trials. Baxter robot efficiently explored the trajectories maintaining a good correlation with the desired velocity, especially in the careful movements, which required lower speeds. The same reasoning was applied for iCub robot, and the resulting correlation coefficients are represented in Figure \ref{fig:icub_pears}. Differently from Baxter, the best results were obtained in the Not Careful, faster movements, with a median correlation above $98\%$. The Careful executed movements were however well correlated with the planned ones, even though ProfileC3 showed a wider distribution. This example is the same whose velocity profiles are reported in Figure \ref{fig:icub_vel_C}, where a delay between the planned and executed movements can be detected in the sagittal and oblique trajectories. This reflects in the Pearson correlation.

\section{Discussion}
In the work we presented, we can identify two main goals: $(i)$ to artificially generate velocity profiles associated to the manipulation of object properties, exploiting generative networks and $(ii)$ to assess whether human-like velocity profiles can be replicated by two different robots. Regarding $(i)$, we proposed a novel conditional GAN architecture that generates new synthetics embeddings, belonging to the Careful and Not Careful class, depending on the provided conditioning label. We showed that our network is able to generate meaningful synthetic profiles, belonging to the same distribution of the original ones without exactly replicating them. Having a network producing artificial samples, instead of simply copying human examples, grants an unlimited source of movements, always different and unique, yet consistent with the object's properties and as communicative as human ones. Moreover, by using a conditional network, we could potentially generate data intermediate to the two initial classes, therefore completely new with respect to the velocity profiles used to train the model. However, the meaning of such intermediate profiles, also in the execution phase, should be properly investigated and assessed in future studies. \\
Concerning $(ii)$, we controlled two distinct robots in performing movements following the selected velocity profiles. From the analyses presented in Sec. \ref{robotresults}, we can generally conclude that both the robots successfully executed the trajectory with the desired velocities. The other considerations are tailored to the two robots, but are meant to discuss what kind of robot characteristics can impact on the synthesis of communicative robotic movements. The amplitude of the executed movements were different between the two robots, due to their dimension, shape and structural design. While all Baxter trajectories covered $60$ $cm$, iCub (a smaller robot) trajectories ranged around $30$ $cm$, with a variability depending on the considered plane. For this reason, even though the original velocity profiles were the same, they were re-scaled differently on the two robots, to cover a different distance with the same duration. Considering the two classes C and NC, we detected a difference in the robots' executions. Indeed, according to Pearson correlation coefficient in Figure \ref{fig:pearson}, Baxter performed better in replicating the careful profiles, with lower accelerations and intensity, while iCub correlation respect to the desired profile was higher for the Not Careful class. Figure \ref{fig:peaks} helps better understand the distinction between the two robots. Baxter presents a strong variability in the maximum velocity reached depending on the three directions of movement and, especially for the NC class, it fails to meet the planned magnitude. iCub shows a more uniform performance respect to the movement directions, however in the NC class we notice again a disparity between executed and planned velocity peaks. This is an important result for future interaction experiments, where the movement directions should be chosen along the most appropriate plan, especially in Baxter case.\\
We posit that the lower performance detected with Baxter robot, can be attributed to a combination of some mechanical limits of its actuators and of MoveIt! controller. Indeed, Baxter is a low-cost manifacturing platform with well known constraints, due to the choice of prioritizing safety over accuracy in its movements \cite{baxter1}. The discrepancy detected on the planes where the movement occurs can be explained by a different involvement of the joints during the action. Such difference is noticeable especially for the NC movements, which required higher accelerations (see Figure \ref{fig:baxter_vel_NC}). Baxter, in the first part of the movement, is not able to keep up with the acceleration of the planned profile, causing a delay of the moment when the maximum peak speed is reached. According to the available information \cite{datasheet}, Baxter is able to reach speeds up to $1$ $m/s$, hence the generated NC profiles are in the correct range, however it may need a longer time to achieve such magnitudes. \\
We can conclude that our approach successfully generates meaningful velocity profiles, and that it is possible to reproduce with different robotic platforms movements belonging either to the C or NC class. However, robots limitations should be taken into account when pursuing a similar approach. The advantage of using MoveIt! to control Baxter robot is that our method can be easily exported to any other robotics platform based on a ROS framework, for further testing with robots with different characteristics. \\
Even though the tested movements were associated with the manipulation of objects with peculiar properties, our approach offers a much broader perspective in the generation of communicative movements. Indeed, the controllers we designed can potentially follow any kind of velocity profile, obviously staying within the motion limits of the robots. For this reason, given the proper velocity input, the end-effector could be controlled to communicate other objects properties, such as the weight, or even an emotional attitude in the gesture, as a gentle or a rude one.

\section{Conclusion}
Within this work, we showed how a generative approach can be used to generalize and autonomously synthesize human-like velocity profiles, associated with careful or not actions. From the control point of view, we proved that is possible to replicate the same kind of movements with the end-effectors of two quite different robotic platforms.
In future studies we plan to assess how the robots movements are perceived by humans interacting with them. In particular, to assess whether is sufficient to modulate the end-effector adopted velocity to communicate the carefulness in the gestures or if other joints in the kinematic chain should be properly controlled. Given the implementation on two quite different robots (iCub, humanoid, and Baxter, collaborative but with an arm kinematic chain far from the human one), their movements, even if controlled with the same principle, could produce very different effect on the interaction. For instance, it could be interesting to assess how the communicative attitude affects the trust and the perceived ability of the two robots.

\section*{Acknowledgments}
\noindent Luca Garello and Linda Lastrico contributed equally to this work, hence they share the first name. \\
The authors would like to warmly thank Alessandro Carf\'i and Simone Macci\'o from DIBRIS, University of Genoa, who guided them in the generation of movements on the Baxter robot.\\
This work has been supported by the project APRIL under the European Union’s Horizon 2020 research and innovation programme, G.A. No 870142.\\
This work has been partially carried out at the Machine Learning Genoa (MaLGa) center, University of Genova, and supported by AFOSR within the project ``Cognitively-inspired architectures for human motion understanding'', grant number FA8655-20-1-7035. This project was also partly conducted at EmaroLab, University of Genova, and supported by the European Commission within the Horizon 2020 Framework (CHIST-ERA, 2014-2020), project InDex.\\
Alessandra Sciutti is supported by a Starting Grant from the European Research Council (ERC) under the European Union’s Horizon 2020 research and innovation programme. G.A. No 804388, wHiSPER.

\bibliographystyle{IEEEtran}
\bibliography{references}

\end{document}